\documentclass[letterpaper, 10 pt, conference]{ieeeconf}  

\IEEEoverridecommandlockouts                              

\usepackage{float}

\usepackage{url}
\usepackage[T1]{fontenc}
\usepackage{newtxtext,newtxmath} 
\usepackage{cite} 
\usepackage{amsmath} 
\usepackage{graphicx}
\usepackage[table]{xcolor}
\usepackage{colortbl}
\usepackage{textcomp}
\usepackage{booktabs}
\usepackage{algorithm}
\usepackage[noend]{algpseudocode}
\usepackage{bm}
\usepackage{makecell}
\usepackage{tikz}
\usetikzlibrary{positioning, arrows.meta, shapes.geometric, calc}
\usepackage[hidelinks]{hyperref} 

\tikzset{
  >=Latex,
  box/.style={
    draw, rounded corners,
    align=center,
    inner sep=3pt, outer sep=0pt,
    minimum width=6.0cm, minimum height=6mm
  },
  boxJ/.style={box, fill=gray!20},
  boxG/.style={box, fill=green!20},
  boxB/.style={box, fill=blue!20},
  edge/.style={->, line width=0.9pt},
  edgeDashed/.style={edge, dashed},
  lab/.style={font=\scriptsize, fill=white, fill opacity=0, text opacity=1, inner sep=1pt}
}

\graphicspath{{figs/}{figures/}{images/}}

\title{\LARGE \bf STORM: Search-Guided Generative World Models for Robotic Manipulation}

\author{Wenjun Lin$^{1}$,Jensen Zhang$^{1}$, Kaitong Cai$^{1}$, Keze Wang$^{1*}$%
\thanks{$^{1}$Sun Yat-sen University, Guangzhou, China.}%
\thanks{*Corresponding author: kezewang@gmail.com}%
}

\begin{document}

\maketitle

\begin{abstract}
We present STORM (Search-Guided Generative World Models), a novel framework for spatio-temporal reasoning that unifies diffusion-based action generation, conditional video prediction, and search-based planning. In contrast to prior Vision-Language-Action (VLA) models that delegate reasoning to language components or abstract latent dynamics, STORM grounds search in explicit visual rollouts, enabling more interpretable and robust long-horizon planning. A diffusion-based VLA module proposes diverse candidate actions, a generative video predictor simulates their outcomes, and Monte Carlo Tree Search (MCTS) selectively refines plans through foresight-driven evaluation.
On the SimplerEnv manipulation benchmark, STORM achieves a new state-of-the-art average success rate of 51.0\%, surpassing CogACT (47.9\%) and other strong baselines. Reward-augmented video prediction improves spatio-temporal fidelity, reducing FVD by over 75\% compared to action-only models. Crucially, STORM’s ability to re-plan after initial failures showcases its advantage over reactive policies. These results demonstrate that search-guided generative world models outperform latent abstractions, establishing a new paradigm for interpretable, foresight-driven decision-making at the intersection of generative modeling and planning.\end{abstract}
\section{Introduction}
\label{sec:intro}

Embodied intelligence is widely regarded as a critical milestone toward Artificial General Intelligence (AGI)\cite{liu2024aligningcyberspacephysical,wang2025embodiedagireviewembodied,jiang2025embodiedintelligencekeyunblocking,duan2022surveyembodiedaisimulators,z1,z2,z3,z19}, aiming to build agents that can perceive, reason, and act in the physical world to accomplish complex tasks such as cooking, tool use, or dexterous object manipulation~\cite{sims2022conceptual,wu2025generativemultiagentcollaborationembodied}. Achieving this vision requires not only accurate multi-modal perception but also robust reasoning and foresight to support high-quality decision-making. In recent years, large-scale \textbf{Vision-Language-Action (VLA)} models---such as OpenVLA~\cite{kim2024openvla}, RT-2~\cite{brohan2023rt2}, and CogACT~\cite{li2024cogactfoundationalvisionlanguageactionmodel}---have emerged as the dominant paradigm, leveraging pre-trained knowledge from web-scale foundation models to generalize across tasks and environments.

Despite their remarkable progress, these models confront a shared architectural bottleneck that limits their fine-grained reasoning capabilities\cite{qu2025surveyefficientreasoninglarge,zhang2025cfvlmcounterfactualvisionlanguagefinetuning,z4,Z5,z6,z11}. The prevalent strategy involves freezing a pre-trained vision backbone and delegating complex reasoning to a Large Language Model (LLM) component\cite{hu2021loralowrankadaptationlarge,z7,z8,z9,z10,z12}. This design, while effective for high-level semantic abstraction, suffers from a fundamental mismatch for physical interaction: it forces a lossy projection of rich, continuous spatio-temporal dynamics onto a discrete, symbolic linguistic manifold. Critical information for manipulation---such as subtle spatial relations, contact dynamics, and the precise causal consequences of actions---becomes ambiguous or is lost entirely during this translation. Consequently, existing VLAs often struggle with the very task-oriented, physically-grounded reasoning required for robust manipulation, leading to factual inaccuracies and suboptimal policies.

To overcome these limitations, we argue for a paradigm shift away from purely abstract linguistic reasoning toward \textbf{visual foresight}—a capability, inspired by human cognition, to mentally simulate and evaluate the spatio-temporal outcomes of potential actions \textit{before} execution. Rather than asking an LLM to reason in language about what \textit{might} happen, we empower the agent to \textit{see} what will happen. This motivates our core principle: \textbf{``predict before you act.''} We posit that grounding planning in explicit, simulated physical futures provides a more robust foundation for decision-making than relying on latent linguistic abstractions.

We instantiate this principle in \textbf{STORM (Search-Guided Generative World Models)}, a novel framework for spatio-temporal reasoning in robotic manipulation. STORM synergistically integrates three components: (i) a diffusion-based VLA policy that proposes a diverse set of candidate action sequences, (ii) a conditional video prediction model, acting as a generative world model, to simulate the visual outcomes of these actions, and (iii) a Monte Carlo Tree Search (MCTS)\cite{MCTS} planner that intelligently explores these simulated futures to identify an optimal long-horizon strategy. Unlike prior world models that plan in abstract latent spaces, STORM grounds its search in \textbf{explicit visual rollouts}, yielding more interpretable, verifiable, and robust reasoning about future trajectories.

We evaluate STORM on a suite of challenging manipulation tasks in the \textbf{SimplerEnv benchmark}~\cite{li2024evaluatingrealworldrobotmanipulation} using a WidowX arm. Our results establish a new state-of-the-art, achieving an \textbf{average task success rate of 51.0\%} and outperforming strong baselines like CogACT, Octo, and OpenVLA. Furthermore, we demonstrate that augmenting the video predictor with reward supervision is critical for learning task-aware dynamics, reducing the \textbf{Fr\'echet Video Distance (FVD\cite{unterthiner2019accurategenerativemodelsvideo}) by over 75\%} and enabling more accurate spatio-temporal rollouts. Beyond aggregate metrics, STORM demonstrates superior resilience, successfully re-planning and recovering from failures where reactive policies become trapped in repetitive error loops.

In summary, our contributions are threefold:
\begin{itemize}
    \item We propose \textbf{STORM}, a novel search-guided framework that integrates a diffusion-based VLA, a generative video world model, and MCTS to enable explicit spatio-temporal reasoning for robotic manipulation.
    \item We demonstrate that a reward-augmented video predictor serves as a highly effective generative world model, substantially improving the fidelity and task-relevance of action-conditioned visual rollouts.
    \item We show empirically that STORM outperforms strong VLA baselines in both task success rates and failure recovery, highlighting the benefits of combining generative foresight with search-based planning.
\end{itemize}

\section{Related Work}
\label{sec:related}

\subsection{Vision-Language-Action Models}
\label{ssec:vla}

Vision-Language-Action (VLA) models, an evolution of Vision-Language Models (VLMs), generate actions based on visual and linguistic inputs. By leveraging the reasoning capabilities of their underlying LLMs, models like OpenVLA \cite{kim2024openvla} and CogACT \cite{li2024cogactfoundationalvisionlanguageactionmodel} exhibit strong generalization and can be fine-tuned for various embodied agents. A key challenge in action generation is that for a given state and instruction, multiple valid action sequences may exist. Modeling this as a deterministic mapping can cause the model to learn an average of trajectories, leading to catastrophic failures \cite{liu2025rdt1bdiffusionfoundationmodel}.

To address this, recent VLAs have adopted diffusion-based architectures for their action decoders \cite{octomodelteam2024octoopensourcegeneralistrobot, liu2025hybridvlacollaborativediffusionautoregression, wen2025diffusionvlageneralizableinterpretablerobot,z17} or as their core backbone \cite{liu2025rdt1bdiffusionfoundationmodel}\cite{wen2025lladavlavisionlanguagediffusion,z15,z16}. The diffusion model's inherent ability to handle multi-modal distributions\cite{ho2020denoisingdiffusionprobabilisticmodels,chi2024diffusionpolicyvisuomotorpolicy,z13,z14} allows it to model the complex action space as a continuous conditional probability distribution\cite{janner2022planningdiffusionflexiblebehavior}. This not only improves action coherence but also provides the multi-solution capability that is a crucial prerequisite for search-based planners like MCTS. Furthermore, many state-of-the-art VLAs employ frozen, pre-trained vision encoders \cite{kim2024openvla, brohan2023rt2}. While this strategy efficiently leverages knowledge from large-scale web datasets, the feature extraction process is not guided by the task instruction, thus limiting the model's capacity for explicit, task-centric spatio-temporal reasoning.

\subsection{Video Prediction as a World Model}
\label{ssec:video_pred}

While large-scale video generation models like Sora \cite{liu2024sorareviewbackgroundtechnology} focus on visual quality and diversity for creative purposes, their application in embodied AI is different. Here, the focus shifts to conditional video prediction, where the primary goal is to generate future frames that strictly adhere to conditioning signals like actions or instructions \cite{wu2024ivideogptinteractivevideogptsscalable, gu2024seerlanguageinstructedvideo, nvidia2025cosmosworldfoundationmodel,z18}. In this context, video prediction models function as learned world models, simulating the dynamics of the environment.

However, language-conditioned video prediction faces significant hurdles. The ambiguity of natural language means that similar instructions can correspond to vastly different motions in training data, making precise control difficult.\cite{fan2025generalizablebimanualfoundationpolicy} In contrast, action-conditioned prediction is more robust. Actions are represented by precise numerical vectors, providing a strong, unambiguous signal. This allows models pre-trained on large web datasets to be effectively fine-tuned on smaller, domain-specific interaction datasets to achieve reliable video prediction, an approach we adopt in our work.

\subsection{Search-based Planning with MCTS}
\label{ssec:mcts}

Search-based planning, particularly Monte Carlo Tree Search (MCTS) combined with a learned dynamics model, is a state-of-the-art paradigm for long-horizon decision-making.\cite{AlphaGo,Schrittwieser_2020,kwak2024efficientmontecarlotree,} This paradigm is exemplified by the seminal work of MuZero \cite{Schrittwieser_2020} and other approaches that plan within a learned, abstract latent space\cite{luo2024pretrainedvisualdynamicsrepresentations,hafner2020dreamcontrollearningbehaviors,sobal2025learningrewardfreeofflinedata}. Our framework, STORM, is inspired by this approach but introduces a critical innovation: we replace the abstract latent model with a generative video predictor that serves as an \textit{explicit, visually-grounded dynamics model}. This allows MCTS to search through concrete, simulated visual futures to select the optimal plan. Crucially, our decoupled architecture treats the VLA as a \textbf{black-box proposal policy}. This makes STORM highly modular, enabling it to be integrated with various existing VLA models without requiring internal modifications or extensive retraining.



\begin{figure*}[t]
    \centering
    \includegraphics[width=\textwidth]{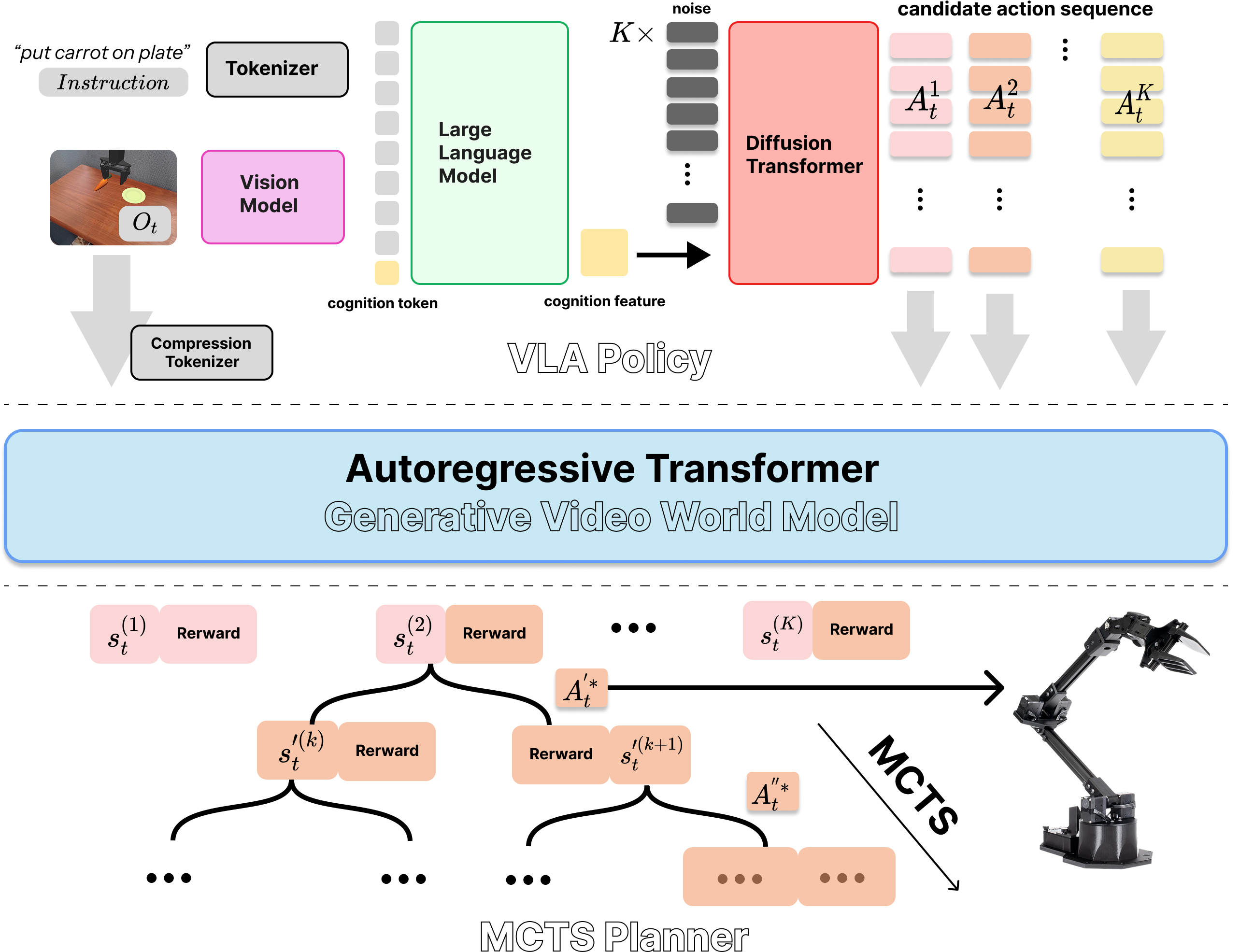}
    \caption{The overall architecture of STORM. The decision loop follows Eq.~\ref{eq:storm_framework}: MCTS orchestrates the process, using the VLA ($\pi_{\text{vla}}$) to propose candidate actions and the video predictor ($M_w$) to simulate their outcomes, ultimately selecting the optimal action $A_t^*$.}
    \label{fig:model_architecture}
\end{figure*}
\section{Methodology}
\label{sec:method}

Our framework, STORM, introduces a generative spatio-temporal reasoning loop for robotic manipulation. Building on this, we first formulate the task as a Partially Observable Markov Decision Process (POMDP), defined by the tuple $(\mathcal{S}, \mathcal{A}, \mathcal{T}, \mathcal{R}, \Omega, \mathcal{O}, \gamma)$, where $\mathcal{S}$ denotes the underlying state space (often unobservable directly), $\mathcal{A}$ is the continuous action space, $\mathcal{T}: \mathcal{S} \times \mathcal{A} \to \mathcal{S}$ represents the probabilistic state transition dynamics, $\mathcal{R}: \mathcal{S} \times \mathcal{A} \to \mathbb{R}$ is the reward function quantifying task progress, $\Omega$ is the observation space (e.g., RGB images), $\mathcal{O}: \mathcal{S} \to \Omega$ maps states to observations, and $\gamma \in [0,1)$ is the discount factor for future rewards. This formulation captures the inherent uncertainty in robotic manipulation, where agents must infer hidden states (e.g., object poses or dynamics) from partial observations and plan long-horizon actions to maximize cumulative rewards.

At each timestep $t$, the agent receives an observation $O_t \in \Omega$ (e.g., camera images) and forms a belief state $s_t$ by integrating $O_t$ with the language instruction $I$ (e.g., via multimodal embeddings). From here, STORM operates as an online planning agent that approximates the optimal action-value function $Q^*(s, a) = \mathbb{E}[\sum_{k=0}^\infty \gamma^k r_{t+k} | s_t = s, a_t = a]$, which represents the expected discounted return following action $a$ in state $s$. This approximation is achieved through lookahead search, enabling deliberate, foresight-driven decision-making that mitigates the limitations of reactive policies. The core decision process instantiates Monte Carlo Tree Search (MCTS), synergistically integrating two generative models: a Vision-Language-Action (VLA) policy for proposing diverse actions ($\pi_{\text{vla}}$) and a video prediction model for simulating outcomes ($M_w$). This integration forms a cohesive loop, where proposals from $\pi_{\text{vla}}$ feed into simulations by $M_w$, which in turn inform the MCTS search. The optimal action sequence $A_t^*$ is thus derived as:
\begin{equation}
A_{t}^{*} = \text{MCTS}(s_{t}, \pi_{\text{vla}}, M_{w}).
\label{eq:storm_framework}
\end{equation}

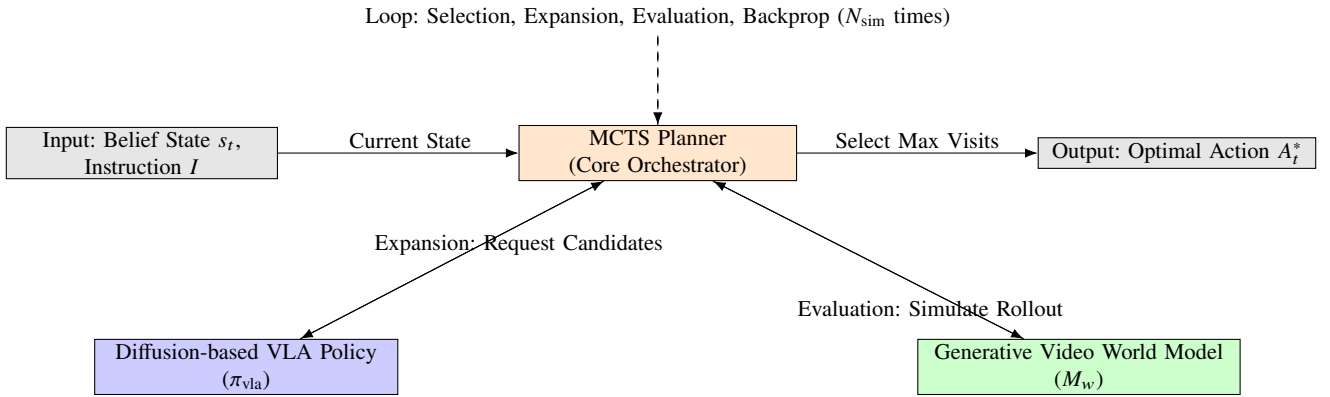
\begin{figure*}[t]
\centering
\begin{tikzpicture}[
  node distance=2.2cm,
  auto,
  every node/.style={scale=0.85, inner sep=2pt},
  >=Latex
]
  \node[draw, rectangle, fill=gray!20, text width=4.1cm, align=center] (input)
    {Input: Belief State $s_t$,\\ Instruction $I$};

  \node[draw, rectangle, fill=orange!20, right=3.2cm of input, text width=4.2cm, align=center] (mcts)
    {MCTS Planner\\ (Core Orchestrator)};

  \node[draw, rectangle, fill=blue!20, below left=2.1cm and 1.6cm of mcts, text width=4.6cm, align=center] (vla)
    {Diffusion-based VLA Policy\\ ($\pi_{\text{vla}}$)};

  \node[draw, rectangle, fill=green!20, below right=2.1cm and 1.6cm of mcts, text width=4.9cm, align=center] (world)
    {Generative Video World Model\\ ($M_w$)};

  \node[draw, rectangle, fill=gray!20, right=3.2cm of mcts, text width=4.2cm, align=center] (output)
    {Output: Optimal Action $A_t^*$};

  \draw[->] (input) -- (mcts) node[pos=0.55, above] {Current State};

  \draw[->] (mcts) -- (vla)
    node[pos=0.28, below, yshift=-2pt, align=center] {Expansion: Request Candidates};

  \draw[->] (mcts) -- (world)
    node[pos=0.70, below, yshift=-2pt, align=center] {Evaluation: Simulate Rollout};

  \draw[->] (vla) -- (mcts);
  \draw[->] (world) -- (mcts);

  \draw[->, dashed] (mcts) -- ++(0,1.6cm) -| (mcts)
    node[midway, above] {Loop: Selection, Expansion, Evaluation, Backprop ($N_{\text{sim}}$ times)};

  \draw[->] (mcts) -- (output) node[midway, above] {Select Max Visits};
\end{tikzpicture}
\caption{Overview of the STORM framework, illustrating the decision loop orchestrated by MCTS. It integrates the VLA policy for proposing diverse action candidates and the video world model for simulating visual outcomes and rewards. The dashed loop represents iterative simulations (selection, expansion, evaluation, backpropagation) for foresight-driven planning. \emph{Note: STORM enables re-planning for failure recovery by grounding search in explicit visual rollouts.}}
\label{fig:storm_overview}
\end{figure*}

\subsection{Action Proposal via Diffusion-based VLA Policy ($\pi_{\text{vla}}$)}
To initiate the planning loop efficiently, search in high-dimensional action spaces must be guided toward promising regions. Building on the POMDP setup, the VLA policy $\pi_{\text{vla}}$ serves as a learned prior, proposing high-probability action candidates conditioned on the multimodal belief state $s_t$. This sets the foundation for subsequent simulation and search steps.

We adopt a diffusion-based architecture for $\pi_{\text{vla}}$, motivated by diffusion models' theoretical strength in modeling complex, multi-modal distributions \cite{sohl2015deep}. Formally, diffusion models learn to reverse a forward noising process that gradually corrupts data into Gaussian noise. The reverse process is parameterized as a denoising network $\epsilon_\theta$, trained to minimize the variational bound on the negative log-likelihood: $\mathbb{E}_{t,\mathbf{x}_0,\epsilon} [\lVert \epsilon - \epsilon_\theta(\sqrt{\bar{\alpha}_t} \mathbf{x}_0 + \sqrt{1-\bar{\alpha}_t} \epsilon, t) \rVert^2]$, where $\bar{\alpha}_t$ controls the noise schedule. At inference, starting from noise $\mathbf{x}_T \sim \mathcal{N}(0, \mathbf{I})$, the model iteratively denoises to sample from $p_\theta(\mathbf{x}_0 | s_t)$, yielding action sequences $A$.

This design is pivotal for manipulation tasks, as it captures action multi-modality—multiple valid trajectories may achieve the same goal (e.g., grasping from different angles). Unlike deterministic policies that regress to mean behaviors and risk mode collapse, diffusion enables sampling diverse, high-likelihood sequences. Consequently, $\pi_{\text{vla}}$ outputs a set of $K$ candidates with associated priors, directly feeding into the MCTS expansion:
\begin{equation}
\pi_{\text{vla}}(s_{t}) \rightarrow \{ (A^{(1)}, p^{(1)}), (A^{(2)}, p^{(2)}), \dots, (A^{(K)}, p^{(K)}) \},
\end{equation}
where $p^{(k)}$ reflects the model's confidence. These proposals prune the search space exponentially, enhancing MCTS efficiency by focusing on task-relevant branches and providing a seamless transition to outcome simulation.

\subsection{Outcome Simulation via Generative Video World Model ($M_w$)}
Following action proposal, the next logical step is to evaluate potential outcomes. The video prediction module $M_w$ acts as a generative world model, approximating the POMDP's transition and reward functions $p(s_{t+1}, r_t | s_t, A_t)$ without explicit physics simulation. It enables visual foresight by simulating "what-if" scenarios based on the proposed actions, bridging the gap between proposal and search-based refinement.

Based on iVideoGPT \cite{wu2024ivideogptinteractivevideogptsscalable}, $M_w$ employs an autoregressive transformer on quantized visual tokens. Observations are encoded into discrete tokens via a Vector-Quantized Variational Autoencoder (VQ-VAE), whose evidence lower bound (ELBO) optimizes reconstruction and codebook commitment: $\mathcal{L}_{\text{VQ}} = \lVert x - \hat{x} \rVert^2 + \lVert \text{sg}[z_e(x)] - e \rVert^2 + \beta \lVert z_e(x) - \text{sg}[e] \rVert^2$. The transformer then predicts future tokens autoregressively, conditioned on past tokens, instruction embeddings (from a frozen LLM), and actions $A^{(j)}$.

To align predictions with task goals, we fine-tune with a hybrid loss:
\begin{equation}
\mathcal{L} = \mathcal{L}_{\text{video}} + \lambda_{\text{reward}}\mathcal{L}_{\text{reward}},
\label{eq:world_model_loss}
\end{equation}
where $\mathcal{L}_{\text{video}}$ is cross-entropy over tokens (ensuring visual fidelity), and $\mathcal{L}_{\text{reward}}$ is MSE on predicted rewards (guiding task-aware dynamics). This reward augmentation theoretically enhances the model's internal representation of causal structures, as ablations show it reduces prediction errors in task-critical dimensions (e.g., object interactions). The output is:
\begin{equation}
M_{w}(s_t, A^{(j)}) \rightarrow (s'_{t}, \hat{r}_t),
\end{equation}
with $s'_t$ as generated frames and $\hat{r}_t$ as a scalar reward estimate, facilitating value-based evaluation in the subsequent MCTS phase.

\subsection{Search-Guided Planning with MCTS}
With actions proposed and outcomes simulated, MCTS orchestrates the entire loop for lookahead planning, building a tree where nodes are states and edges are actions with statistics $\{N(s,A), W(s,A), Q(s,A), P(s,A)\}$ (visit count, total value, mean value, prior). It performs $N_{\text{sim}}$ simulations per decision (Algorithm \ref{alg:storm}), balancing exploration and exploitation via Upper Confidence Bound (UCB) algorithm \cite{6145622}, and closes the loop by selecting the optimal action for execution.

\begin{enumerate}
\small
    \item \textbf{Selection:} From root $s_t$, recursively choose actions maximizing the PUCT score:
    \begin{equation}
    A = \arg\max_{A'} \left( Q(s, A') + c_{\text{puct}} \cdot P(s, A') \cdot \frac{\sqrt{\sum_B N(s, B)}}{1 + N(s, A')} \right)
    \label{eq:puct}
    \end{equation}
    derived from UCB, ensuring asymptotic optimality under bandit assumptions.
    
    \item \textbf{Expansion:} At leaf $s_L$, invoke $\pi_{\text{vla}}(s_L)$ to generate $K$ actions, initializing child nodes with priors $P(s_L, A^{(k)}) = p^{(k)}$.
    
    \item \textbf{Evaluation:} Simulate one child using $M_w(s_L, A^{(k)}) \to (s', \hat{r})$, yielding value $V = \hat{r}$ (or discounted if multi-step).
    
    \item \textbf{Backpropagation:} Update path statistics:
    \begin{align*}
        N(s,A) &\leftarrow N(s,A) + 1, \\
        W(s,A) &\leftarrow W(s,A) + V, \\
        Q(s,A) &\leftarrow W(s,A) / N(s,A).
    \end{align*}
\end{enumerate}
Post-simulations, select $A_t^* = \arg\max_A N(\text{root}, A)$ for robustness, enabling re-planning and failure recovery.

\begin{algorithm}[h!]
    \normalsize 
    \caption{STORM Decision-Making Process}
    \label{alg:storm}
    \begin{algorithmic}[1]
        \Require Current state $s_t$, VLA policy $\pi_{vla}$, World Model $M_w$, Simulations $N_{sim}$
        \State Initialize tree $T$ with root $s_t$
        \For{$i=1$ to $N_{sim}$}
            \State \textbf{Selection:} node $\gets$ T.root
            \While{node not leaf}
                \State action $\gets$ \Call{SelectChild}{node} \Comment{PUCT (Eq.~\ref{eq:puct})}
                \State node $\gets$ T.GetChild(node, action)
            \EndWhile
            \If{node not expanded}
                \State (actions, priors) $\gets \pi_{vla}$(node.state)
                \ForAll{(a,p) in (actions,priors)} \State T.Expand(node,a,p) \EndFor
            \EndIf
            \State \textbf{Evaluation:} child $\gets$ unvisited child of node
            \State (next, r) $\gets M_w$(node.state, child.action)
            \State child.state $\gets$ next; value $\gets r
            $
            \State \textbf{Backpropagation:} \While{node $\neq$ null} \State \Call{UpdateStats}{node,value}; node $\gets$ node.parent \EndWhile
        \EndFor
        \State \Return \Call{SelectActionWithMaxVisits}{T.root}
    \end{algorithmic}
\end{algorithm}
\section{Experiments}
\label{sec:experiments}

\begin{figure*}[t]
    \centering
    \includegraphics[width=\textwidth]{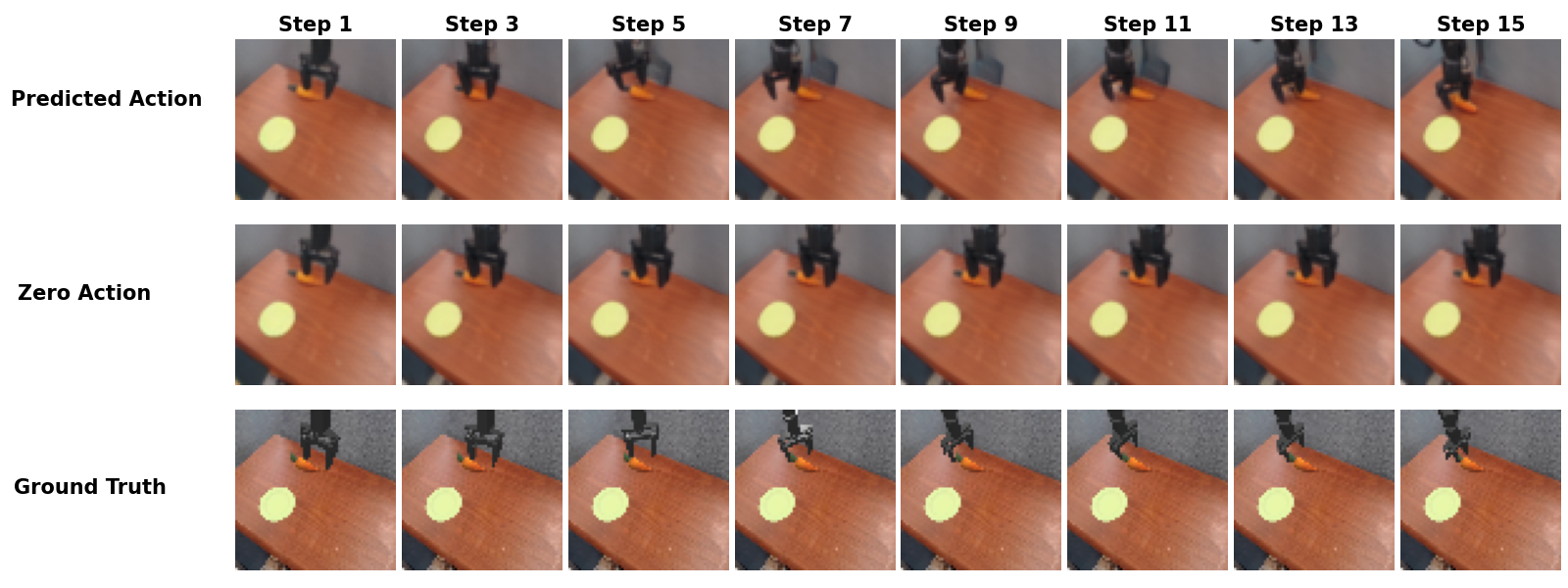}
    \caption{Qualitative results for video prediction (Task: Put Carrot on Plate). The prediction conditioned on the VLA's action (middle) aligns well with the ground truth (bottom).}
    \label{fig:action-conditional-predict-spoon}
\end{figure*}

\begin{table*}[t]
    \centering
    \caption{Success rates (\%) on SimplerEnv manipulation tasks. \textbf{STORM} achieves the best performance across all tasks, outperforming prior state-of-the-art methods including its base model \textbf{CogACT}.}
    \label{tab:widowx_simpler}
    \resizebox{\textwidth}{!}{
    \begin{tabular}{lccccc}
        \toprule
        \textbf{Method} & 
        \textbf{Put Spoon on Towel} & 
        \textbf{Put Carrot on Plate} &
        \textbf{Stack Green on Yellow Block} &
        \textbf{Put Eggplant in Basket} &
        \textbf{Average} \\
        \midrule
        RT-1-X~\cite{embodimentcollaboration2024openxembodimentroboticlearning} & 0.0 & 4.2 & 0.0 & 0.0 & 1.1 \\
        Octo-Base~\cite{octomodelteam2024octoopensourcegeneralistrobot} & 15.8 & 12.5 & 0.0 & 41.7 & 17.5 \\
        Octo-Small~\cite{octomodelteam2024octoopensourcegeneralistrobot} & 41.7 & 8.2 & 0.0 & 56.7 & 26.7 \\
        OpenVLA~\cite{kim2024openvla} & 4.2 & 0.0 & 0.0 & 12.5 & 4.2 \\
        CogACT~\cite{li2024cogactfoundationalvisionlanguageactionmodel} & \textbf{75.0} & \textbf{45.9} & \textbf{12.5} & \textbf{58.3} & \textbf{47.9} \\
        \rowcolor{gray!10}
        \textbf{Ours (STORM)} & \textbf{79.2} & \textbf{50.0} & \textbf{12.5} & \textbf{62.5} & \textbf{51.0} \\
        \bottomrule
    \end{tabular}
    }
\end{table*}

\subsection{Experimental Setup}
\label{ssec:setup}

\textbf{Environment and Data.} We evaluate our framework on robotic manipulation tasks using the WidowX robot arm in the SimplerEnv simulator \cite{li2024evaluatingrealworldrobotmanipulation}, which shows a strong correlation with real-world performance. The primary training data for our modules is the Bridge dataset \cite{ebert2021bridgedataboostinggeneralization}.

\textbf{Models and Training.} Our VLA module is the pre-trained \textbf{CogACT-Base (7B parameters)} \cite{li2024cogactfoundationalvisionlanguageactionmodel}, used without further training. Our generative world model is based on the pre-trained \textbf{iVideoGPT-medium} Transformer \cite{wu2024ivideogptinteractivevideogptsscalable}. We fine-tune this model on the Bridge dataset to introduce action-conditioning and a reward prediction head. The fine-tuning was performed on \textbf{two NVIDIA A100 (80GB) GPUs}, training for approximately \textbf{120,000 steps}. The per-device batch size was \textbf{18} (total batch size of 36) with no gradient accumulation. We used the \textbf{AdamW optimizer} with a learning rate of \textbf{5e-4}, a \textbf{cosine learning rate schedule}, and a weight decay of \textbf{0.01}. To ensure training stability, we employed \textbf{gradient clipping} with a maximum norm of 30.0. The training was conducted \textbf{without mixed-precision}. The reward loss weight $\lambda_{\text{reward}}$ was set to 20.

\textbf{MCTS Parameters.} In our planner, we use $N_{\text{sim}}=8$ simulations per decision step, with a planning depth of $D=3$. The discount factor $\gamma$ is 0.9, and the exploration constant $c_{\text{puct}}$ is 1.0. The VLA proposes $K=8$ candidate actions at each expansion step.

\textbf{Evaluation Metrics.} For the main tasks, we report the average success rate over 24 procedural variations per task. For the video prediction ablation, we use standard metrics including Fréchet Video Distance (FVD) \cite{unterthiner2019accurategenerativemodelsvideo}, LPIPS \cite{zhang2018unreasonableeffectivenessdeepfeatures}, PSNR \cite{huynh-thu2008psnr}, and SSIM \cite{WANG2004ssim}, visualized in a radar chart for comprehensive comparison.

\begin{figure*}[t]
    \centering
    \includegraphics[width=\textwidth]{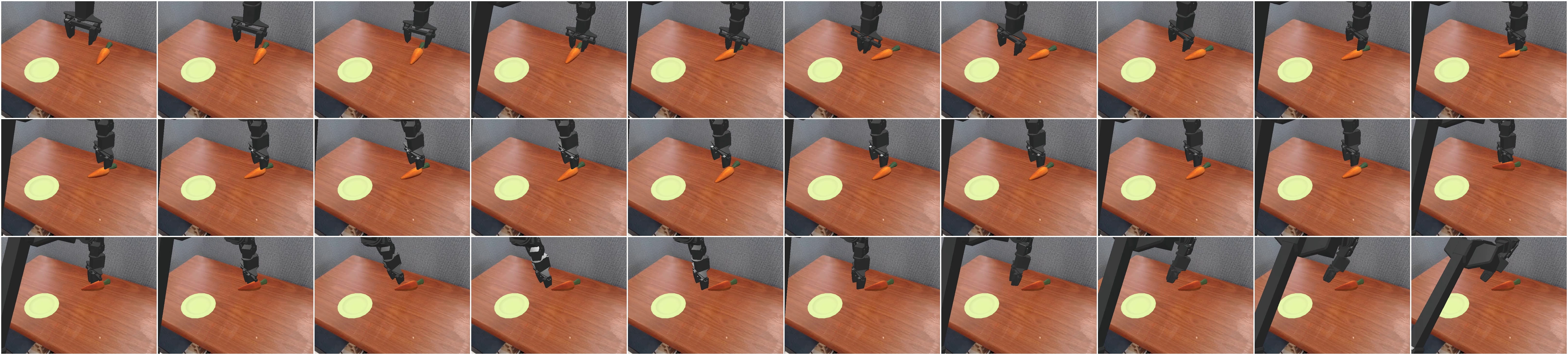}
    \vspace{2mm} 
    \includegraphics[width=\textwidth]{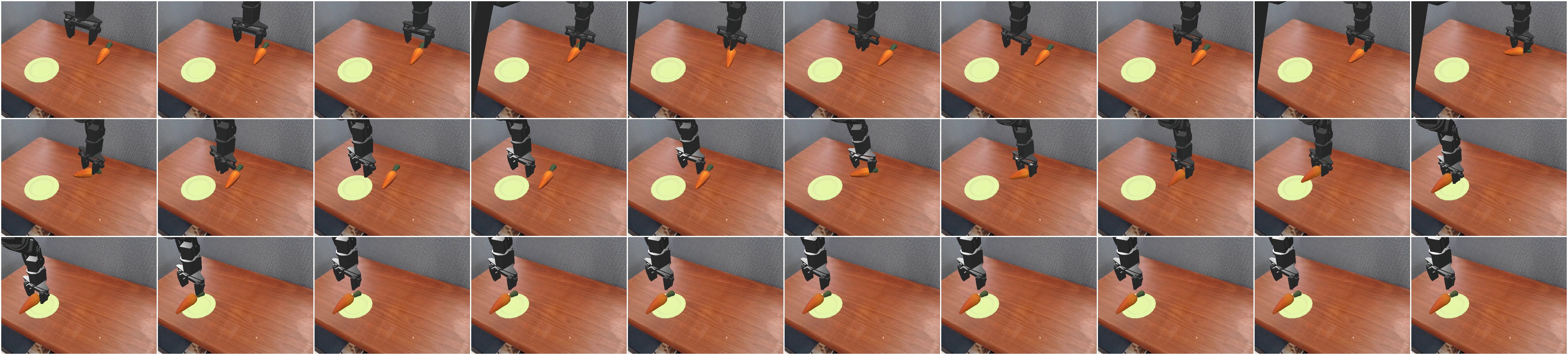}
    \caption{Case study on the ``Put Carrot on Plate'' task, demonstrating STORM's ability to recover from failure. \textbf{Top}: The baseline CogACT model fails, getting stuck in a repetitive loop after initial unsuccessful grasp attempts. \textbf{Bottom}: Our model, STORM, uses its lookahead planning to re-evaluate after the same initial failures and finds a new, successful trajectory to complete the task.}
    \label{fig:failure-vs-success}
\end{figure*}

\subsection{Qualitative Analysis: Visual Foresight in Action}

To qualitatively validate our generative world model as a reliable proxy for true environment dynamics, we compare its simulated rollouts against ground-truth executions. As depicted in Fig.~\ref{fig:action-conditional-predict-spoon}, the model demonstrates potent \textbf{visual foresight}. Given a candidate action sequence from the VLA, its predicted future (middle row) accurately captures the salient causal dynamics of the actual execution (bottom row). This result confirms its ability to generate physically plausible, action-conditioned futures, which is the foundational requirement for its use as a simulation engine within MCTS.

Crucially, the model's objective is not pixel-perfect replication but rather the modeling of \textbf{task-relevant causal dynamics}. This principled abstraction is vital for computational tractability, as it allows the planner to efficiently evaluate the consequences of actions without being burdened by irrelevant visual details. This validates the model's design and fitness for purpose.
\subsection{Manipulation Task Performance}

We evaluate STORM's end-to-end performance on four challenging manipulation tasks in SimplerEnv. As detailed in Table~\ref{tab:widowx_simpler}, STORM establishes a new state-of-the-art with an average success rate of 51.0\%, achieving a notable 3.1 percentage point improvement over the highly competitive CogACT baseline (47.9\%) and surpassing all other prior methods. This margin, while numerically modest, is significant in the context of robotic manipulation and points to a fundamental architectural advantage. It provides strong empirical evidence that \textbf{foresight-driven planning with a generative world model is a superior paradigm to purely reactive policies}. 

The core of this advantage lies in overcoming the inherent brittleness of direct state-to-action mapping. Reactive policies can be "myopic," committing to a greedily selected action that may lead to an unrecoverable state. In contrast, STORM's search-based method performs \textbf{combinatorial exploration of future trajectories} in simulation. This allows it to evaluate the long-term consequences of entire action sequences, effectively pruning branches that appear promising locally but are globally suboptimal. This deliberative process results in more robust and successful task execution, particularly in scenarios with complex causal chains.ba

The performance on the ``Stack Green Block on Yellow Block'' task, where STORM matches the baseline, is also highly insightful. This task's success is dominated by the precise control of contact-rich physics—a domain where visual prediction models can struggle to capture the subtle, non-linear dynamics with perfect fidelity. This result suggests that the performance of our search-guided \textit{framework} is ultimately bounded by the predictive fidelity of its current world model \textit{instantiation}. This is not a failure of the search paradigm itself, but rather a clear and promising direction for future work: integrating more physically-informed or higher-fidelity generative models into the STORM framework will directly unlock superior performance on the most delicate and precise manipulation tasks.

\subsection{Case Study: Planning for Recovery}

The principal advantage of search-based planning over reactive control is the ability to reason through complex situations and execute strategic recovery from failure. Figure~\ref{fig:failure-vs-success} provides a compelling case study of this capability. The baseline CogACT model, acting reactively, becomes trapped in a \textbf{policy local minimum}. After two failed grasp attempts, its learned state--action mapping undergoes \textbf{mode collapse}; it repeatedly outputs a high-confidence but incorrect action, leading to a deterministic and irrecoverable failure loop. This behavior highlights a structural limitation: the reactive pipeline provides no mechanism to re-evaluate or revise confident but flawed predictions once committed.

In stark contrast, STORM leverages its closed-loop planning cycle to actively \emph{interrogate} initial errors. The first, failing action suggested by the diffusion-based VLA is simply one branch in the Monte Carlo Tree Search (MCTS). After being rolled out in the reward-augmented world model, this branch is correctly assigned a low value. Guided by the Upper Confidence Bound (UCB) rule, MCTS naturally shifts attention toward unexplored trajectories, balancing exploitation of promising nodes with exploration of alternatives. Crucially, the reward-aware world model provides task-centric causal feedback—capturing object interactions and goal configurations rather than superficial pixel similarity—which ensures that branches leading to recovery trajectories are preferentially selected. This iterative evaluation–backpropagation loop allows STORM to escape local minima that are intractable for reactive policies. which ensures that branches leading to recovery trajectories are preferentially selected. This iterative evaluation–backpropagation loop allows STORM to escape local minima that are intractable for reactive policies, effectively embodying a \emph{fail-and-adapt} principle where early errors become signals for globally successful re-planning.
\begin{figure}[t]
    \centering
    \includegraphics[width=\linewidth]{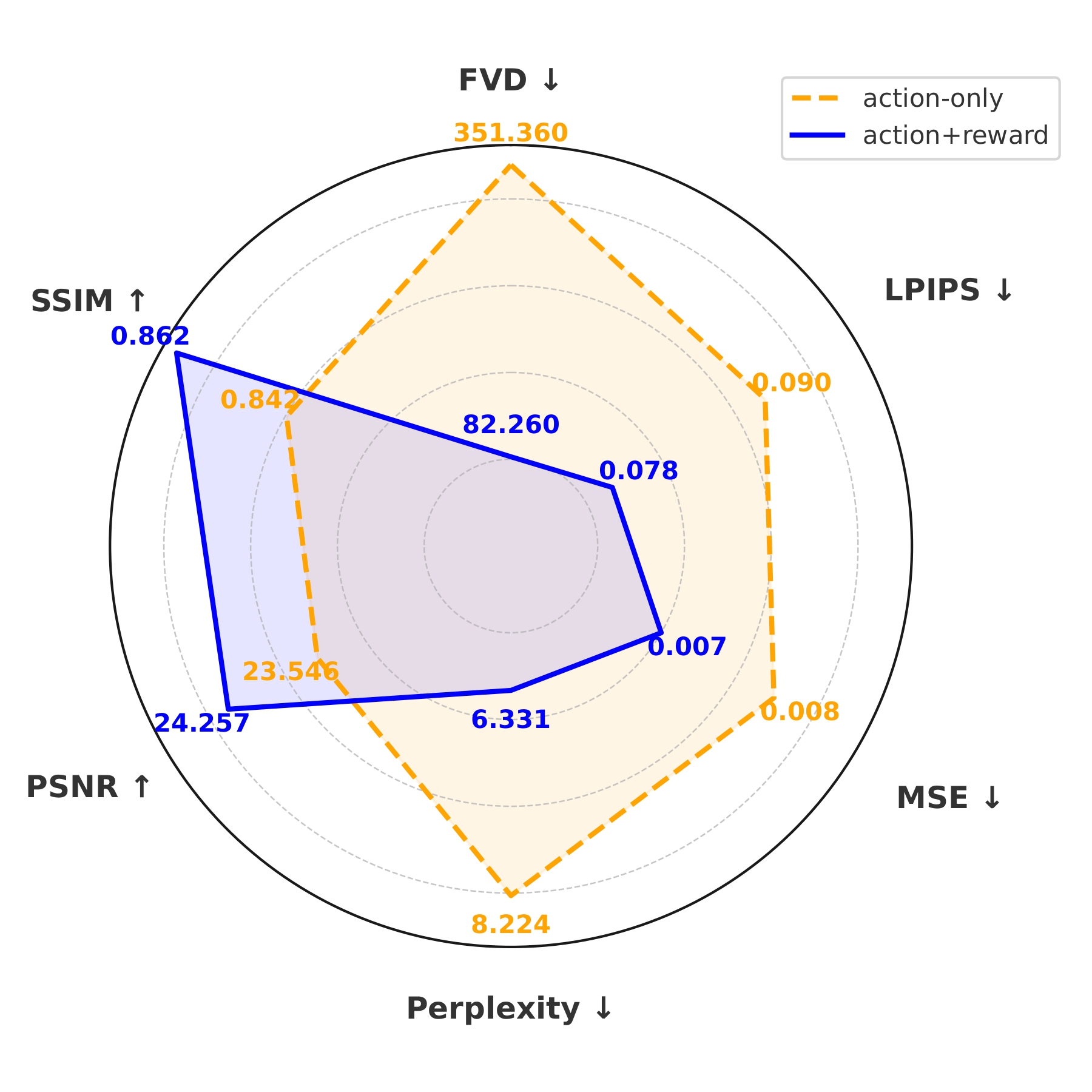}
    \caption{\textbf{Reward supervision is critical for learning a high-fidelity world model.} The radar chart compares video prediction metrics for a model trained with our full objective (`action+reward') versus one without reward supervision (`action-only'). The reward-augmented model's superior performance across all axes demonstrates that this signal compels the model to learn \textbf{task-relevant causal structures}, moving beyond superficial visual patterns to enable effective, foresight-driven planning.}
    \label{fig:quantitative_results}
\end{figure}

\subsection{Ablation Study: Impact of Reward Prediction}
\label{ssec:ablation}

While essential, traditional component-removal ablations are uninformative for a deeply integrated, closed-loop system like STORM, where the VLA ($\pi_{\text{vla}}$), the world model ($M_w$), and MCTS form a synergistic loop. Removing any component fundamentally changes the algorithm's class; for instance, a "VLA-only" configuration is simply the reactive CogACT baseline.

Therefore, we argue that the most insightful ablation examines the \textbf{quality of the information exchanged within the loop}, which is the primary bottleneck for effective planning. We test the hypothesis that reward supervision is critical for learning task-relevant dynamics by comparing our full model (`action+reward') against a version trained without the reward prediction head (`action-only').

The results in Figure~\ref{fig:quantitative_results} are decisive. The `action+reward' model significantly outperforms its counterpart across all metrics. This empirically validates our central theoretical claim: the reward signal acts as a powerful \textbf{information bottleneck} that regularizes the high-dimensional, under-constrained problem of video prediction. Without this signal, the model's loss function incentivizes capturing any statistically predictable pattern, including visually salient but causally irrelevant details (e.g., background textures). The inclusion of the $\mathcal{L}_{\text{reward}}$ term, through a shared network backbone, forces the model's latent representations to be \textbf{disentangled}, prioritizing the \textbf{task-relevant causal structures}—contact dynamics, object displacement, goal configurations—as these are the sole predictors of future reward.

From the perspective of our POMDP formulation, this is crucial. The `action-only` model learns a proxy for the transition dynamics $p(s_{t+1} | s_t, A_t)$, while our `action+reward` model learns to approximate the \textit{joint} distribution $p(s_{t+1}, r_t | s_t, A_t)$. This is theoretically superior because the objective of the planner (maximizing cumulative reward) is now directly reflected in the world model's training objective. The substantial improvement in Fréchet Video Distance (FVD) confirms this: the model generates a more realistic \textit{distribution} of future trajectories, which is paramount for the statistical validity of Monte Carlo search. Concurrently, better SSIM/PSNR scores indicate a more accurate prediction of the spatial arrangements of objects—the geometric precondition for task success. In essence, this ablation confirms that an effective world model for planning must be more than a generative simulator; it must be a \textbf{value-aware simulator}. This task-centric, causal understanding is what empowers STORM's deliberative planning and is a key contribution of our work.

\section{Conclusion}
\label{sec:conclusion}

In this paper, we introduced STORM, a novel framework for generative spatio-temporal reasoning in robotic manipulation. We address the limitations of VLA models that rely on frozen visual encoders and purely linguistic reasoning. By integrating a diffusion-based VLA with a generative video world model and an MCTS planner, STORM explicitly simulates and evaluates multiple future trajectories proposed by the VLA, enabling it to select a more optimal action plan. Our experiments demonstrate that this approach improves the average task success rate by 3.1 percentage points over a strong CogACT baseline \cite{li2024cogactfoundationalvisionlanguageactionmodel}, validating the benefit of generative visual planning.
its promising performance, STORM has limitations. The computational cost of the video model restricts the depth and breadth of the MCTS search, and its predictive precision can be insufficient for highly fine-grained manipulation tasks. Our future work will focus on two primary directions. First, we plan to explore joint training, using the action values derived from the MCTS search as an auxiliary supervisory signal to fine-tune the VLA policy. This could instill the VLA with an implicit planning capability. Second, to enhance planning efficiency, we will investigate distilling the video world model, trading pixel-level fidelity for faster simulation speeds, thereby enabling deeper and more effective searches.

\newpage
\bibliographystyle{IEEEtran}

\bibliography{reference}

@misc{liu2024aligningcyberspacephysical,
  author        = {Yang Liu and Weixing Chen },
  title         = {Aligning Cyber Space with Physical World: A Comprehensive Survey on Embodied AI},
  year          = {2024},
  eprint        = {2407.06886},
  archivePrefix = {arXiv},
  primaryClass  = {cs.CV},
  note          = {[EB/OL] [2025-05-06]},
  url           = {https://arxiv.org/abs/2407.06886}
}

@misc{hu2021loralowrankadaptationlarge,
      title={LoRA: Low-Rank Adaptation of Large Language Models}, 
      author={Edward J. Hu and Yelong Shen and Phillip Wallis },
      year={2021},
      eprint={2106.09685},
      archivePrefix={arXiv},
      primaryClass={cs.CL},
      url={https://arxiv.org/abs/2106.09685}, 
}

@misc{brohan2023rt2,
      title={RT-2: Vision-Language-Action Models Transfer Web Knowledge to Robotic Control}, 
      author={Anthony Brohan and Noah Brown and Justice Carbajal},
      year={2023},
      eprint={2307.15818},
      archivePrefix={arXiv},
      primaryClass={cs.RO},
      url={https://arxiv.org/abs/2307.15818}, 
}

@misc{kim2024openvla,
      title={OpenVLA: An Open-Source Vision-Language-Action Model}, 
      author={Moo Jin Kim and Karl Pertsch and Siddharth Karamcheti and Ted Xiao and Ashwin Balakrishna and Suraj Nair },
      year={2024},
      eprint={2406.09246},
      archivePrefix={arXiv},
      primaryClass={cs.RO},
      url={https://arxiv.org/abs/2406.09246}, 
}

@article{sims2022conceptual,
  author    = {Chris R. Sims and Rachel A. Lerch and John A. Tarduno and Robert A. Jacobs},
  title     = {Conceptual knowledge shapes visual working memory for complex visual information},
  journal   = {Scientific Reports},
  year      = {2022},
  volume    = {12},
  number    = {1},
  pages     = {8088},
  doi       = {10.1038/s41598-022-12137-0},
  url       = {https://doi.org/10.1038/s41598-022-12137-0}
}

@misc{liu2025rdt1bdiffusionfoundationmodel,
      title={RDT-1B: a Diffusion Foundation Model for Bimanual Manipulation}, 
      author={Songming Liu and Lingxuan Wu and Bangguo Li},
      year={2025},
      eprint={2410.07864},
      archivePrefix={arXiv},
      primaryClass={cs.RO},
      url={https://arxiv.org/abs/2410.07864}, 
}

@misc{octomodelteam2024octoopensourcegeneralistrobot,
      title={Octo: An Open-Source Generalist Robot Policy}, 
      author={Octo Model Team and Dibya Ghosh and Homer Walke },
      year={2024},
      eprint={2405.12213},
      archivePrefix={arXiv},
      primaryClass={cs.RO},
      url={https://arxiv.org/abs/2405.12213}, 
}

@misc{li2024cogactfoundationalvisionlanguageactionmodel,
      title={CogACT: A Foundational Vision-Language-Action Model for Synergizing Cognition and Action in Robotic Manipulation}, 
      author={Qixiu Li and Yaobo Liang },
      year={2024},
      eprint={2411.19650},
      archivePrefix={arXiv},
      primaryClass={cs.RO},
      url={https://arxiv.org/abs/2411.19650}, 
}

@misc{liu2024sorareviewbackgroundtechnology,
      title={Sora: A Review on Background, Technology, Limitations, and Opportunities of Large Vision Models}, 
      author={Yixin Liu and Kai Zhang and Yuan Li and Zhiling Yan and Chujie Gao },
      year={2024},
      eprint={2402.17177},
      archivePrefix={arXiv},
      primaryClass={cs.CV},
      url={https://arxiv.org/abs/2402.17177}, 
}

@misc{wu2024ivideogptinteractivevideogptsscalable,
      title={iVideoGPT: Interactive VideoGPTs are Scalable World Models}, 
      author={Jialong Wu and Shaofeng Yin and Ningya Feng and Xu He and Dong Li },
      year={2024},
      eprint={2405.15223},
      archivePrefix={arXiv},
      primaryClass={cs.CV},
      url={https://arxiv.org/abs/2405.15223}, 
}

@misc{gu2024seerlanguageinstructedvideo,
      title={Seer: Language Instructed Video Prediction with Latent Diffusion Models}, 
      author={Xianfan Gu and Chuan Wen and Weirui Ye and Jiaming Song and Yang Gao},
      year={2024},
      eprint={2303.14897},
      archivePrefix={arXiv},
      primaryClass={cs.CV},
      url={https://arxiv.org/abs/2303.14897}, 
}

@misc{nvidia2025cosmosworldfoundationmodel,
      title={Cosmos World Foundation Model Platform for Physical AI}, 
      author={NVIDIA and Niket Agarwal and Arslan Ali and Maciej Bala },
      year={2025},
      eprint={2501.03575},
      archivePrefix={arXiv},
      primaryClass={cs.CV},
      url={https://arxiv.org/abs/2501.03575}, 
}

@article{6145622,
  author={Browne, Cameron B. and Powley, Edward and Whitehouse},
  journal={IEEE Transactions on Computational Intelligence and AI in Games}, 
  title={A Survey of Monte Carlo Tree Search Methods}, 
  year={2012},
  volume={4},
  number={1},
  pages={1-43},
  doi={10.1109/TCIAIG.2012.2186810}}

@article{Schrittwieser_2020,
   title={Mastering Atari, Go, chess and shogi by planning with a learned model},
   volume={588},
   ISSN={1476-4687},
   url={http://dx.doi.org/10.1038/s41586-020-03051-4},
   DOI={10.1038/s41586-020-03051-4},
   number={7839},
   journal={Nature},
   publisher={Springer Science and Business Media LLC},
   author={Schrittwieser, Julian and Antonoglou, Ioannis and Hubert, Thomas and Simonyan, Karen and Sifre, Laurent and Schmitt, Simon and Guez, Arthur and Lockhart, Edward and Hassabis, Demis and Graepel, Thore and Lillicrap, Timothy and Silver, David},
   year={2020},
   month=dec, pages={604–609} }

@misc{li2024evaluatingrealworldrobotmanipulation,
      title={Evaluating Real-World Robot Manipulation Policies in Simulation}, 
      author={Xuanlin Li and Kyle Hsu and Jiayuan Gu},
      year={2024},
      eprint={2405.05941},
      archivePrefix={arXiv},
      primaryClass={cs.RO},
      url={https://arxiv.org/abs/2405.05941}, 
}

@misc{ebert2021bridgedataboostinggeneralization,
      title={Bridge Data: Boosting Generalization of Robotic Skills with Cross-Domain Datasets}, 
      author={Frederik Ebert and Yanlai Yang and Karl Schmeckpeper and Bernadette Bucher and Georgios Georgakis and Kostas Daniilidis and Chelsea Finn and Sergey Levine},
      year={2021},
      eprint={2109.13396},
      archivePrefix={arXiv},
      primaryClass={cs.RO},
      url={https://arxiv.org/abs/2109.13396}, 
}

@misc{embodimentcollaboration2024openxembodimentroboticlearning,
      title={Open X-Embodiment: Robotic Learning Datasets and RT-X Models}, 
      author={Embodiment Collaboration and Abby O'Neill and Abdul Rehman and Abhinav Gupta },
      year={2024},
      eprint={2310.08864},
      archivePrefix={arXiv},
      primaryClass={cs.RO},
      url={https://arxiv.org/abs/2310.08864}, 
}

@misc{unterthiner2019accurategenerativemodelsvideo,
      title={Towards Accurate Generative Models of Video: A New Metric \& Challenges}, 
      author={Thomas Unterthiner and Sjoerd van Steenkiste and Karol Kurach and Raphael Marinier and Marcin Michalski and Sylvain Gelly},
      year={2019},
      eprint={1812.01717},
      archivePrefix={arXiv},
      primaryClass={cs.CV},
      url={https://arxiv.org/abs/1812.01717}, 
}

@article{huynh-thu2008psnr,  
  author = {Quan Huynh-Thu and Mohammed Ghanbari},  
  title = {Scope of validity of {PSNR} in image/video quality assessment},  
  journal = {Electronics Letters},  
  volume = {44},  
  number = {13},  
  pages = {800--801},  
  year = {2008},  
  publisher = {IET}  
}

@article{WANG2004ssim,
author = {Wang, Zhou and Bovik, Alan and Sheikh, Hamid and Simoncelli, Eero},
year = {2004},
month = {05},
pages = {600 - 612},
title = {Image Quality Assessment: From Error Visibility to Structural Similarity},
volume = {13},
journal = {Image Processing, IEEE Transactions on},
doi = {10.1109/TIP.2003.819861}
}

@misc{zhang2018unreasonableeffectivenessdeepfeatures,
      title={The Unreasonable Effectiveness of Deep Features as a Perceptual Metric}, 
      author={Richard Zhang and Phillip Isola and Alexei A. Efros and Eli Shechtman and Oliver Wang},
      year={2018},
      eprint={1801.03924},
      archivePrefix={arXiv},
      primaryClass={cs.CV},
      url={https://arxiv.org/abs/1801.03924}, 
}

@inproceedings{MCTS,
author = {Kocsis, Levente and Szepesvári, Csaba},
year = {2006},
month = {09},
pages = {282-293},
title = {Bandit Based Monte-Carlo Planning},
volume = {2006},
isbn = {978-3-540-45375-8},
journal = {Machine Learning: ECML},
doi = {10.1007/11871842_29}
}

@article{AlphaGo,
  author    = {David Silver and Aja Huang and Chris J. Maddison and Arthur Guez and Laurent Sifre and George van den Driessche and Julian Schrittwieser and Ioannis Antonoglou and Veda Panneershelvam and Marc Lanctot and Sander Dieleman and Dominik Grewe and John Nham and Nal Kalchbrenner and Ilya Sutskever and others},
  title     = {Mastering the game of Go with deep neural networks and tree search},
  journal   = {Nature},
  volume    = {529},
  number    = {7587},
  pages     = {484--489},
  year      = {2016},
  doi       = {10.1038/nature16961},
  url       = {https://doi.org/10.1038/nature16961},
  issn      = {1476-4687}
}

@misc{liu2025hybridvlacollaborativediffusionautoregression,
      title={HybridVLA: Collaborative Diffusion and Autoregression in a Unified Vision-Language-Action Model}, 
      author={Jiaming Liu and Hao Chen and Pengju An and Zhuoyang Liu and Renrui Zhang and Chenyang Gu and Xiaoqi Li and Ziyu Guo and Sixiang Chen and Mengzhen Liu and Chengkai Hou and Mengdi Zhao and KC alex Zhou and Pheng-Ann Heng and Shanghang Zhang},
      year={2025},
      eprint={2503.10631},
      archivePrefix={arXiv},
      primaryClass={cs.CV},
      url={https://arxiv.org/abs/2503.10631}, 
}

@misc{wen2025lladavlavisionlanguagediffusion,
      title={LLaDA-VLA: Vision Language Diffusion Action Models}, 
      author={Yuqing Wen and Hebei Li and Kefan Gu and Yucheng Zhao and Tiancai Wang and Xiaoyan Sun},
      year={2025},
      eprint={2509.06932},
      archivePrefix={arXiv},
      primaryClass={cs.RO},
      url={https://arxiv.org/abs/2509.06932}, 
}

@misc{wen2025diffusionvlageneralizableinterpretablerobot,
      title={Diffusion-VLA: Generalizable and Interpretable Robot Foundation Model via Self-Generated Reasoning}, 
      author={Junjie Wen and Minjie Zhu and Yichen Zhu and Zhibin Tang and Jinming Li and Zhongyi Zhou and Chengmeng Li and Xiaoyu Liu and Yaxin Peng and Chaomin Shen and Feifei Feng},
      year={2025},
      eprint={2412.03293},
      archivePrefix={arXiv},
      primaryClass={cs.RO},
      url={https://arxiv.org/abs/2412.03293}, 
}

@misc{fan2025generalizablebimanualfoundationpolicy,
      title={Towards a Generalizable Bimanual Foundation Policy via Flow-based Video Prediction}, 
      author={Chenyou Fan and Fangzheng Yan and Chenjia Bai and Jiepeng Wang and Chi Zhang and Zhen Wang and Xuelong Li},
      year={2025},
      eprint={2505.24156},
      archivePrefix={arXiv},
      primaryClass={cs.CV},
      url={https://arxiv.org/abs/2505.24156}, 
}

@misc{kwak2024efficientmontecarlotree,
      title={Efficient Monte Carlo Tree Search via On-the-Fly State-Conditioned Action Abstraction}, 
      author={Yunhyeok Kwak and Inwoo Hwang and Dooyoung Kim and Sanghack Lee and Byoung-Tak Zhang},
      year={2024},
      eprint={2406.00614},
      archivePrefix={arXiv},
      primaryClass={cs.LG},
      url={https://arxiv.org/abs/2406.00614}, 
}

@misc{luo2024pretrainedvisualdynamicsrepresentations,
      title={Pre-trained Visual Dynamics Representations for Efficient Policy Learning}, 
      author={Hao Luo and Bohan Zhou and Zongqing Lu},
      year={2024},
      eprint={2411.03169},
      archivePrefix={arXiv},
      primaryClass={cs.CV},
      url={https://arxiv.org/abs/2411.03169}, 
}

@misc{hafner2020dreamcontrollearningbehaviors,
      title={Dream to Control: Learning Behaviors by Latent Imagination}, 
      author={Danijar Hafner and Timothy Lillicrap and Jimmy Ba and Mohammad Norouzi},
      year={2020},
      eprint={1912.01603},
      archivePrefix={arXiv},
      primaryClass={cs.LG},
      url={https://arxiv.org/abs/1912.01603}, 
}

@misc{sobal2025learningrewardfreeofflinedata,
      title={Learning from Reward-Free Offline Data: A Case for Planning with Latent Dynamics Models}, 
      author={Vlad Sobal and Wancong Zhang and Kyunghyun Cho and Randall Balestriero and Tim G. J. Rudner and Yann LeCun},
      year={2025},
      eprint={2502.14819},
      archivePrefix={arXiv},
      primaryClass={cs.LG},
      url={https://arxiv.org/abs/2502.14819}, 
}

@misc{wu2025generativemultiagentcollaborationembodied,
      title={Generative Multi-Agent Collaboration in Embodied AI: A Systematic Review}, 
      author={Di Wu and Xian Wei and Guang Chen and Hao Shen and Xiangfeng Wang and Wenhao Li and Bo Jin},
      year={2025},
      eprint={2502.11518},
      archivePrefix={arXiv},
      primaryClass={cs.MA},
      url={https://arxiv.org/abs/2502.11518}, 
}

@misc{zhang2025cfvlmcounterfactualvisionlanguagefinetuning,
      title={CF-VLM:CounterFactual Vision-Language Fine-tuning}, 
      author={Jusheng Zhang and Kaitong Cai and Yijia Fan and Jian Wang and Keze Wang},
      year={2025},
      eprint={2506.17267},
      archivePrefix={arXiv},
      primaryClass={cs.LG},
      url={https://arxiv.org/abs/2506.17267}, 
}

@misc{qu2025surveyefficientreasoninglarge,
      title={A Survey of Efficient Reasoning for Large Reasoning Models: Language, Multimodality, and Beyond}, 
      author={Xiaoye Qu and Yafu Li and Zhaochen Su and Weigao Sun and Jianhao Yan and Dongrui Liu and Ganqu Cui and Daizong Liu and Shuxian Liang and Junxian He and Peng Li and Wei Wei and Jing Shao and Chaochao Lu and Yue Zhang and Xian-Sheng Hua and Bowen Zhou and Yu Cheng},
      year={2025},
      eprint={2503.21614},
      archivePrefix={arXiv},
      primaryClass={cs.CL},
      url={https://arxiv.org/abs/2503.21614}, 
}

@misc{wang2025embodiedagireviewembodied,
      title={Toward Embodied AGI: A Review of Embodied AI and the Road Ahead}, 
      author={Yequan Wang and Aixin Sun},
      year={2025},
      eprint={2505.14235},
      archivePrefix={arXiv},
      primaryClass={cs.AI},
      url={https://arxiv.org/abs/2505.14235}, 
}

@misc{jiang2025embodiedintelligencekeyunblocking,
      title={Embodied Intelligence: The Key to Unblocking Generalized Artificial Intelligence}, 
      author={Jinhao Jiang and Changlin Chen and Shile Feng and Wanru Geng and Zesheng Zhou and Ni Wang and Shuai Li and Feng-Qi Cui and Erbao Dong},
      year={2025},
      eprint={2505.06897},
      archivePrefix={arXiv},
      primaryClass={cs.AI},
      url={https://arxiv.org/abs/2505.06897}, 
}

@misc{duan2022surveyembodiedaisimulators,
      title={A Survey of Embodied AI: From Simulators to Research Tasks}, 
      author={Jiafei Duan and Samson Yu and Hui Li Tan and Hongyuan Zhu and Cheston Tan},
      year={2022},
      eprint={2103.04918},
      archivePrefix={arXiv},
      primaryClass={cs.AI},
      url={https://arxiv.org/abs/2103.04918}, 
}

@misc{janner2022planningdiffusionflexiblebehavior,
      title={Planning with Diffusion for Flexible Behavior Synthesis}, 
      author={Michael Janner and Yilun Du and Joshua B. Tenenbaum and Sergey Levine},
      year={2022},
      eprint={2205.09991},
      archivePrefix={arXiv},
      primaryClass={cs.LG},
      url={https://arxiv.org/abs/2205.09991}, 
}

@misc{chi2024diffusionpolicyvisuomotorpolicy,
      title={Diffusion Policy: Visuomotor Policy Learning via Action Diffusion}, 
      author={Cheng Chi and Zhenjia Xu and Siyuan Feng and Eric Cousineau and Yilun Du and Benjamin Burchfiel and Russ Tedrake and Shuran Song},
      year={2024},
      eprint={2303.04137},
      archivePrefix={arXiv},
      primaryClass={cs.RO},
      url={https://arxiv.org/abs/2303.04137}, 
}

@misc{ho2020denoisingdiffusionprobabilisticmodels,
      title={Denoising Diffusion Probabilistic Models}, 
      author={Jonathan Ho and Ajay Jain and Pieter Abbeel},
      year={2020},
      eprint={2006.11239},
      archivePrefix={arXiv},
      primaryClass={cs.LG},
      url={https://arxiv.org/abs/2006.11239}, 
}

@inproceedings{
z1,
title={{KABB}: Knowledge-Aware Bayesian Bandits for Dynamic Expert Coordination in Multi-Agent Systems},
author={Jusheng Zhang and Zimeng Huang and Yijia Fan and Ningyuan Liu and Mingyan Li and Zhuojie Yang and Jiawei Yao and Jian Wang and Keze Wang},
booktitle={Forty-second International Conference on Machine Learning},
year={2025},
url={https://openreview.net/forum?id=AKvy9a4jho}
}

@inproceedings{
z2,
title={{GAM}-Agent: Game-Theoretic and Uncertainty-Aware Collaboration for Complex Visual Reasoning},
author={Jusheng Zhang and Yijia Fan and Wenjun Lin and Ruiqi Chen and Haoyi Jiang and Wenhao Chai and Jian Wang and Keze Wang},
booktitle={The Thirty-ninth Annual Conference on Neural Information Processing Systems},
year={2025},
url={https://openreview.net/forum?id=EKJhU5ioSo}
}

@misc{z3,
      title={CF-VLM:CounterFactual Vision-Language Fine-tuning}, 
      author={Jusheng Zhang and Kaitong Cai and Yijia Fan and Jian Wang and Keze Wang},
      year={2025},
      eprint={2506.17267},
      archivePrefix={arXiv},
      primaryClass={cs.LG},
      url={https://arxiv.org/abs/2506.17267}, 
}

@inproceedings{
z4,
title={{MAT}-Agent: Adaptive Multi-Agent Training Optimization},
author={Jusheng Zhang and Kaitong Cai and Yijia Fan and Ningyuan Liu and Keze Wang},
booktitle={The Thirty-ninth Annual Conference on Neural Information Processing Systems},
year={2025},
url={https://openreview.net/forum?id=YDWRTYgR79}
}

@inproceedings{
Z5,
title={Tri-{MARF}: A Tri-Modal Multi-Agent Responsive Framework for Comprehensive 3D Object Annotation},
author={Jusheng Zhang and Yijia Fan and Zimo Wen and Jian Wang and Keze Wang},
booktitle={The Thirty-ninth Annual Conference on Neural Information Processing Systems},
year={2025},
url={https://openreview.net/forum?id=YGIbwfNWot}
}

@misc{z6,
      title={MM-CoT:A Benchmark for Probing Visual Chain-of-Thought Reasoning in Multimodal Models}, 
      author={Jusheng Zhang and Kaitong Cai and Xiaoyang Guo and Sidi Liu and Qinhan Lv and Ruiqi Chen and Jing Yang and Yijia Fan and Xiaofei Sun and Jian Wang and Ziliang Chen and Liang Lin and Keze Wang},
      year={2025},
      eprint={2512.08228},
      archivePrefix={arXiv},
      primaryClass={cs.CV},
      url={https://arxiv.org/abs/2512.08228}, 
}

@misc{z7,
      title={HybridToken-VLM: Hybrid Token Compression for Vision-Language Models}, 
      author={Jusheng Zhang and Xiaoyang Guo and Kaitong Cai and Qinhan Lv and Yijia Fan and Wenhao Chai and Jian Wang and Keze Wang},
      year={2025},
      eprint={2512.08240},
      archivePrefix={arXiv},
      primaryClass={cs.CV},
      url={https://arxiv.org/abs/2512.08240}, 
}

@misc{z8,
      title={Failure-Driven Workflow Refinement}, 
      author={Jusheng Zhang and Kaitong Cai and Qinglin Zeng and Ningyuan Liu and Stephen Fan and Ziliang Chen and Keze Wang},
      year={2025},
      eprint={2510.10035},
      archivePrefix={arXiv},
      primaryClass={cs.AI},
      url={https://arxiv.org/abs/2510.10035}, 
}

@misc{z9,
      title={OSC: Cognitive Orchestration through Dynamic Knowledge Alignment in Multi-Agent LLM Collaboration}, 
      author={Jusheng Zhang and Yijia Fan and Kaitong Cai and Xiaofei Sun and Keze Wang},
      year={2025},
      eprint={2509.04876},
      archivePrefix={arXiv},
      primaryClass={cs.AI},
      url={https://arxiv.org/abs/2509.04876}, 
}

@misc{z10,
      title={Learning Dynamics of VLM Finetuning}, 
      author={Jusheng Zhang and Kaitong Cai and Jing Yang and Keze Wang},
      year={2025},
      eprint={2510.11978},
      archivePrefix={arXiv},
      primaryClass={cs.LG},
      url={https://arxiv.org/abs/2510.11978}, 
}

@misc{z11,
      title={DrDiff: Dynamic Routing Diffusion with Hierarchical Attention for Breaking the Efficiency-Quality Trade-off}, 
      author={Jusheng Zhang and Yijia Fan and Kaitong Cai and Zimeng Huang and Xiaofei Sun and Jian Wang and Chengpei Tang and Keze Wang},
      year={2025},
      eprint={2509.02785},
      archivePrefix={arXiv},
      primaryClass={cs.CL},
      url={https://arxiv.org/abs/2509.02785}, 
}

@misc{z12,
      title={Kolmogorov-Arnold Fourier Networks}, 
      author={Jusheng Zhang and Yijia Fan and Kaitong Cai and Keze Wang},
      year={2025},
      eprint={2502.06018},
      archivePrefix={arXiv},
      primaryClass={cs.LG},
      url={https://arxiv.org/abs/2502.06018}, 
}

@misc{z13,
      title={Top-Down Semantic Refinement for Image Captioning}, 
      author={Jusheng Zhang and Kaitong Cai and Jing Yang and Jian Wang and Chengpei Tang and Keze Wang},
      year={2025},
      eprint={2510.22391},
      archivePrefix={arXiv},
      primaryClass={cs.CV},
      url={https://arxiv.org/abs/2510.22391}, 
}

@misc{z14,
      title={DepthSSC: Monocular 3D Semantic Scene Completion via Depth-Spatial Alignment and Voxel Adaptation}, 
      author={Jiawei Yao and Jusheng Zhang and Xiaochao Pan and Tong Wu and Canran Xiao},
      year={2024},
      eprint={2311.17084},
      archivePrefix={arXiv},
      primaryClass={cs.CV},
      url={https://arxiv.org/abs/2311.17084}, 
}

@inproceedings{z15,
    title = "{CCG}: Rare-Label Prediction via Neural {SEM}{--}Driven Causal Game",
    author = "Fan, Yijia  and
      Zhang, Jusheng  and
      Cai, Kaitong  and
      Yang, Jing  and
      Wang, Keze",
    editor = "Christodoulopoulos, Christos  and
      Chakraborty, Tanmoy  and
      Rose, Carolyn  and
      Peng, Violet",
    booktitle = "Findings of the Association for Computational Linguistics: EMNLP 2025",
    month = nov,
    year = "2025",
    address = "Suzhou, China",
    publisher = "Association for Computational Linguistics",
    url = "https://aclanthology.org/2025.findings-emnlp.331/",
    doi = "10.18653/v1/2025.findings-emnlp.331",
    pages = "6243--6256",
    ISBN = "979-8-89176-335-7"
}

@misc{z16,
      title={3DAlign-DAER: Dynamic Attention Policy and Efficient Retrieval Strategy for Fine-grained 3D-Text Alignment at Scale}, 
      author={Yijia Fan and Jusheng Zhang and Kaitong Cai and Jing Yang and Jian Wang and Keze Wang},
      year={2025},
      eprint={2511.13211},
      archivePrefix={arXiv},
      primaryClass={cs.CV},
      url={https://arxiv.org/abs/2511.13211}, 
}

@misc{z17,
      title={Cost-Effective Communication: An Auction-based Method for Language Agent Interaction}, 
      author={Yijia Fan and Jusheng Zhang and Kaitong Cai and Jing Yang and Chengpei Tang and Jian Wang and Keze Wang},
      year={2025},
      eprint={2511.13193},
      archivePrefix={arXiv},
      primaryClass={cs.AI},
      url={https://arxiv.org/abs/2511.13193}, 
}

@misc{z18,
      title={RaCoT: Plug-and-Play Contrastive Example Generation Mechanism for Enhanced LLM Reasoning Reliability}, 
      author={Kaitong Cai and Jusheng Zhang and Yijia Fan and Jing Yang and Keze Wang},
      year={2025},
      eprint={2510.22710},
      archivePrefix={arXiv},
      primaryClass={cs.AI},
      url={https://arxiv.org/abs/2510.22710}, 
}

@article{z19,
author = {Li, Xiaohua and Zhang, Jusheng and Safara, Fatemeh},
title = {Improving the Accuracy of Diabetes Diagnosis Applications through a Hybrid Feature Selection Algorithm},
year = {2021},
issue_date = {Feb 2023},
publisher = {Kluwer Academic Publishers},
address = {USA},
volume = {55},
number = {1},
issn = {1370-4621},
url = {https://doi.org/10.1007/s11063-021-10491-0},
doi = {10.1007/s11063-021-10491-0},
journal = {Neural Process. Lett.},
month = mar,
pages = {153–169},
numpages = {17}
}

\end{document}